%% file: paper.tex
\documentclass[letterpaper, 10 pt, conference]{ieeeconf}
\IEEEoverridecommandlockouts    
\input{stachnisslab-latex}


\usepackage{booktabs,multirow,siunitx,xcolor,colortbl,subcaption,array}
\usepackage{calc}   
\newcolumntype{I}{!{\hspace{0.7em}\vrule\hspace{0.7em}}}
\newcommand{\met}[2]{\makebox[\widthof{HE~[deg]}+0.4em][l]{#1}\,\textcolor{black}{$#2$}}
\newcommand{\bak}[1]{}

\title{\LARGE \bf FlowRow: Learning Navigation Behaviors in Agricultural Environments with Long-Horizon Goal-Conditioned Flow Matching Policies}

\author{Utkarsh Bajpai \and C\'{e}dric Pradalier \and St\'{e}phanie Aravecchia
  \thanks{All authors are with Georgia Tech-CNRS IRL2958.}%
  \thanks{This work has been funded by the European Union’s Horizon Europe research and innovation programme under the Marie Skłodowska-Curie Actions Doctoral Networks (MSCA-DN), under grant agreement ID 101169330 -- AIGreenBots.
  }%
}

\begin{document}
\thispagestyle{empty}
\pagestyle{empty}
\maketitle

\begin{abstract}

Autonomous monitoring and intervention in semi-structured agricultural environments, such as orchards and vineyards, require mobile robots to navigate to locations across diverse field configurations. This entails performing navigation behaviors including in-row driving, row turning and skipping, and inter-row traversal. Recent learning-based approaches have enabled robust in-row navigation from onboard perception, but navigation across rows typically requires a priori knowledge of field geometry to specify and tune row-transition maneuvers, preventing arbitrary transitions between rows. We address this challenge by framing navigation in these environments as a goal-directed navigation task, where a long-horizon goal is expressed relative to the robot and a planner generates local paths based on the goal and the robot's perceptual observations. We present \textit{FlowRow}, a flow-matching-based local planner which models the distribution of diverse agricultural navigation behaviors as $\mathrm{SE}(2)$ paths using the long-horizon goal and depth perception. To enable systematic evaluation, we introduce \textit{RowNavBench}, a navigation benchmark for semi-structured agricultural environments developed in Isaac Sim, comprising 900 episodes across grape vineyards, lemon orchards, and apple orchards with varying row lengths, row and plant spacings, plant positions, and weed growth. FlowRow achieves 87\% success, compared to 44\% for the strongest generative behavior cloning baseline retrained on the same demonstrations. We will release our benchmark, data, and code upon acceptance.

\end{abstract}

\section{Introduction}
\label{sec:intro}

\begin{figure}[t]
  \centering
  \includegraphics[width=\linewidth]{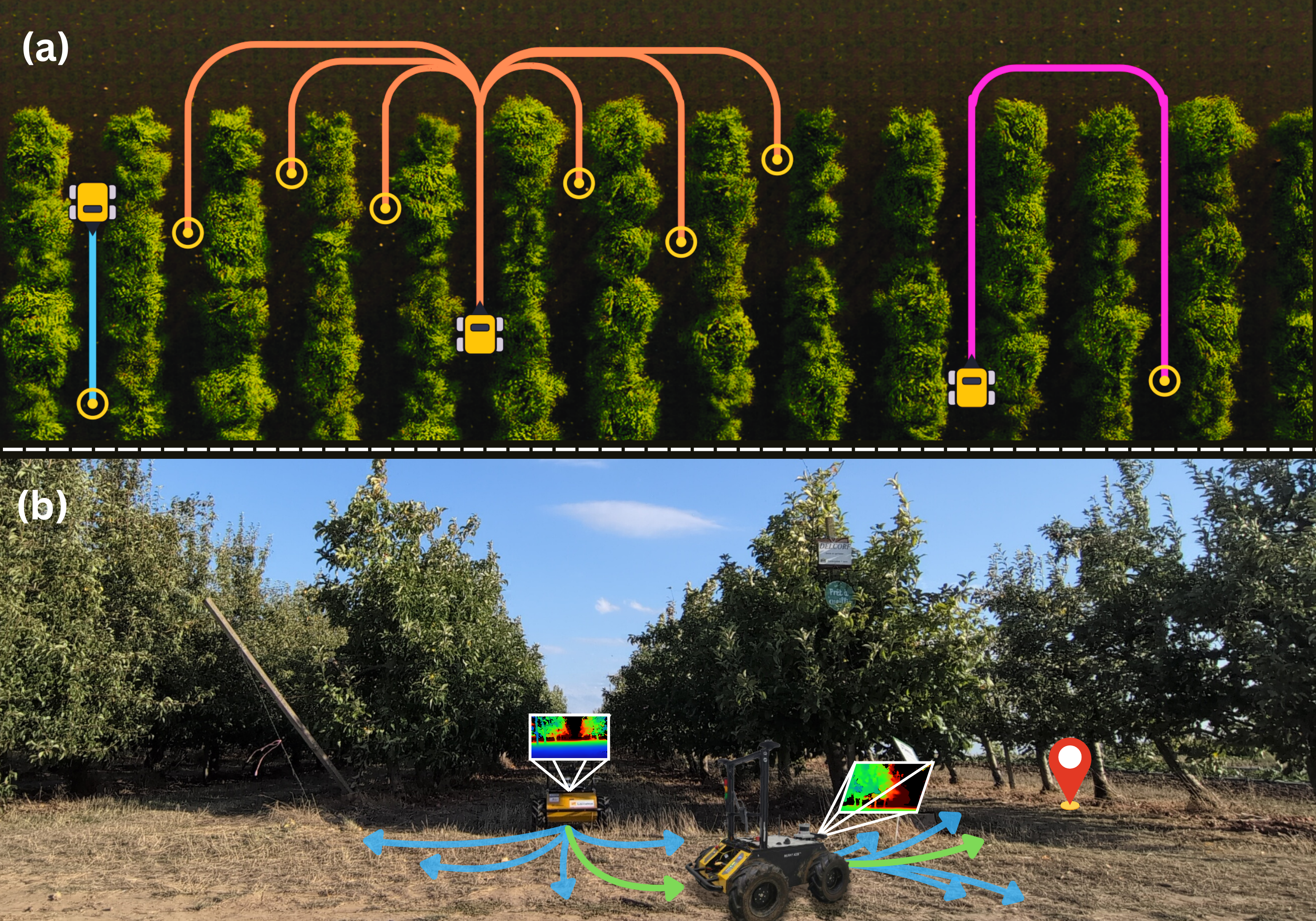}
  \caption{(\textbf{a}) Long-horizon navigation in semi-structured row-crop environments requires combining multiple behaviors, including in-row driving~(blue), headland turning and row skipping~(orange) to traverse between rows~(pink). (\textbf{b}) FlowRow addresses this setting with a flow-matching policy that performs goal-directed navigation from egocentric depth observations. Given a goal (red pin), FlowRow generates candidate local $\mathrm{SE}(2)$ paths (blue) and follows the path that progresses toward the goal (green).}
  \label{fig:motivation}
    \vspace{-0.65cm}
\end{figure}

Autonomous mobile robots are increasingly deployed in agricultural fields for precision intervention tasks such as crop inspection, and harvesting, which require reliable navigation to target locations. Such robots operate in semi-structured environments such as orchards, vineyards, and greenhouses, where crops are arranged in rows separated by navigable corridors with headlands at their ends. Large fields require long-horizon navigation, while crop rows constrain how the robot can traverse the environment. A target location may be spatially close to the robot yet lie in another row, requiring the robot to first move away from it, transition through a headland, and approach it from a different direction. The appropriate trajectory therefore depends not only on the target location, but also on the surrounding row structure. This gives rise to distinct navigation behaviors, including in-row driving, headland turning, and row skipping, as shown in~\figref{fig:motivation}~(a). Agricultural environments also vary considerably in vegetation density and obstacle distribution. Vegetation growth can narrow corridors or alter their apparent structure, while rocks, weeds, and other obstacles can reduce free space. Dense canopies can attenuate or obstruct GNSS signals, and uneven terrain or loose soil can induce wheel slippage and odometry drift. Together, these factors make long-horizon agricultural navigation challenging.

In this work, we consider the problem of goal-directed navigation in semi-structured agricultural environments. Specifically, we consider a robot initialized in an unknown row-crop field with headlands at the ends of the rows. The robot has no prior map or knowledge of field geometry, such as crop and row spacing, but can perceive its surroundings through onboard perception. Given a relative goal position, the robot must navigate to it while avoiding obstacles and performing navigation behaviors, including in-row driving, headland turning, row skipping, or combinations thereof.

Existing agricultural navigation systems often rely on tracking pre-computed reference paths generated from known field geometry \cite{oksanen2009coverage}. While effective, these approaches can degrade from localization errors under canopy, and environmental variation. Recent perception-based methods reduce dependence on localization by using onboard sensing for localization-free in-row navigation~\cite{ahmadi2020visual,sivakumar2021rss}. 
Such methods estimate the row centerline for row following, while row transitions are handled separately through rule-based maneuvers that assume known row geometry~\cite{bertoglio2023ecmr}. 
As a result, these methods do not support navigation to arbitrary goal positions and are limited to turning into the adjacent rows without row skipping or traversing to non-adjacent rows. In contrast, we unify these navigation behaviors through a single goal-conditioned policy which performs local planning driven by egocentric perception, as illustrated in Fig~\ref{fig:motivation}~(b).

The main contributions of this paper are:

\begin{itemize}
\item We introduce \emph{FlowRow}, a flow-matching policy trained entirely in simulation for long-horizon, goal-directed local planning in unknown, semi-structured agricultural environments. FlowRow requires no prior knowledge of environment geometry and supports in-row navigation, row turning, row skipping, and inter-row transitions.

\item We introduce \textit{RowNavBench}, a simulation benchmark comprising 900 episodes across variations in row length, row spacing, row curvature, and vegetation conditions. We evaluate FlowRow on RowNavBench against depth-based local planning baselines, demonstrating over 40\% higher success across the benchmark.

\item We evaluate FlowRow under localization errors against an oracle that tracks a pre-planned global reference path, and show that local planning with FlowRow is more robust to localization errors and results in fewer navigation failures. We further validate FlowRow with a real-world zero-shot deployment.
\end{itemize}

\section{Related Work}
\label{sec:related}

Autonomous navigation in agricultural fields has been studied extensively~\cite{oksanen2009coverage,sivakumar2021rss,ahmadi2020visual}. Conventional navigation systems require the robot to navigate to target locations by following a global reference path defined as a sequence of GNSS waypoints using prior knowledge of the field geometry. This approach is effective when field structure is known and localization remains reliable. However, GNSS performance can degrade under dense canopy and in environments affected by multipath, introducing localization errors that affect reference-path tracking.

\subsection{Perception-based Navigation in Agricultural Fields}
To reduce dependence on global localization and reference-path tracking, several methods use onboard perception for agricultural navigation.
Most works focus on row-following, where the objective is to drive in-row without navigating to a given goal position. Ahmadi~\etalcite{ahmadi2020visual} demonstrate visual-servoing-based row following with a straddler robot by tracking crop lines from a top-down view over small row crops. De Silva~\etalcite{desilva2024jfr} propose a learning-based method for crop-line estimation from similar top-down imagery. Sivakumar~\etalcite{sivakumar2021rss,sivakumar2024rss} instead demonstrate row-following in maize fields from RGB images by predicting crop-row keypoints and tracking the inferred row center with controller. Similarly, Kim~\etalcite{kim2025pagnav} estimate the row center from LiDAR range images and track it with a controller. Headland turns are typically handled separately using state machines or rule-based maneuvers with geometric priors. A common strategy is to exit the current row, execute a $90^\circ$ turn, advance by a distance determined from known field geometry such as row spacing, and turn $90^\circ$ again before resuming row following. Liu~\etalcite{liu2023visioniros} and Ahmadi~\etalcite{ahmadi2020visual} use such predefined maneuvers together with side-facing cameras to determine alignment during the turn. Similarly, Kim~\etalcite{kim2025pagnav} use intensity changes in LiDAR range images to determine the robot's alignment with the rows during the maneuver. Ahmadi~\etalcite{ahmadi2022towards} extend their visual-servoing approach to an omnidirectional straddler platform that performs multi-row switching by counting visible crop rows in the top-down view and driving across them. Prior perception-based approaches have primarily focused on in-row driving, with headland transitions handled through a combination of onboard sensing and predefined maneuvers based on field-geometry priors. To the best of our knowledge, prior work has not demonstrated a local planner that jointly supports in-row navigation, row skipping, and headland transitions from egocentric observations without such priors.

\subsection{Navigation with Imitation Learning}
Imitation learning provides an alternative to designing separate navigation behaviors by learning action-generation policies from demonstrations. Diffusion~\cite{chi2023diffusionpolicy} and flow matching policies~\cite{lipman2023fm} are generative formulations for modeling multimodal action distributions. Diffusion learns through iterative denoising, while flow matching learns a continuous vector field that transports samples from noise to the demonstrated trajectory distribution. In navigation, NoMaD~\cite{sridhar2023nomad}, a diffusion policy, and FlowNav~\cite{gode2025flownav}, a flow-matching policy, navigate toward image-goals using previously acquired topological maps in structured indoor and outdoor environments. NavDP~\cite{cai2025navdp} considers target-location navigation from RGB-D observations, while SanD-Planner~\cite{wang2026sandplanner} encodes goals in cartesian space and performs local planning from depth observations by generating multiple diffusion-based trajectory candidates as $\mathbb{R}^3$ splines. However, transferring such policies to agricultural environments remains challenging. CropFollow++~\cite{sivakumar2024rss} shows that NoMaD can fail to identify the crop corridor under canopy, while the same authors' later diffusion-based study focuses only on agricultural row turning~\cite{sivakumar2024learningturndiffusionimitation} and reports difficulties with in-row control and varied initial conditions. These results suggest that generative navigation policies do not transfer trivially to row-crop environments, and that learning an isolated turning behavior does not address long-horizon goal-directed navigation. We instead consider a single local planner that must infer and compose in-row driving, headland turning, and row-skipping behaviors from egocentric observations and a target location. We therefore introduce \textit{FlowRow}, a flow-matching policy for long-horizon local planning in unknown agricultural environments without explicit field-geometry priors.

\begin{figure*}[t]
  \centering
  \includegraphics[width=\linewidth]{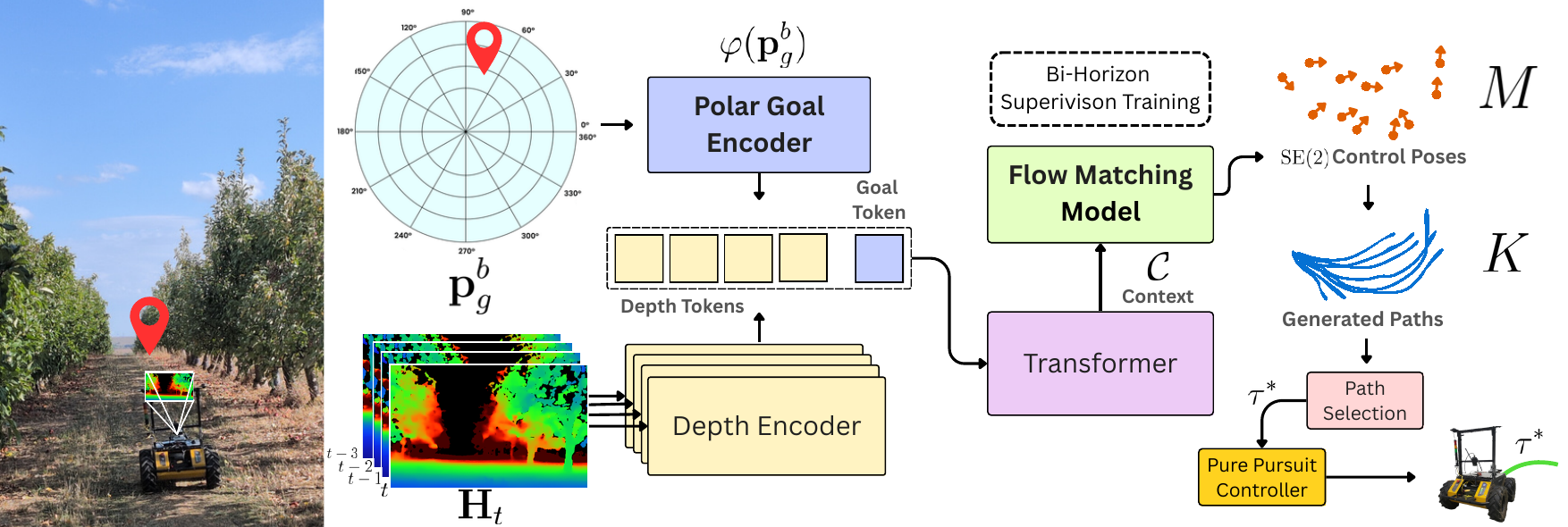}
  \caption{\textbf{Our Approach.} At each inference step, FlowRow encodes a long-horizon goal $\mathbf{p}^b_g$ expressed in the robot's body frame via our polar goal encoder alongwith a history of 4 most recent depth images into a transformer to encode the context vector for the flow matching module. Our flow matching module is trained with trained with Bi-Horizon supervision and generates $32 \times 12$ $\mathrm{SE(2)}$ control poses, which are used to compute 32 candidate paths. A path selection mechanism ranks these paths to select the path to be driven by the robot.}
  \label{fig:approach}
    \vspace{-0.65cm}
\end{figure*}

\section{Problem Formulation}
\label{sec:problem}
We consider a wheeled mobile robot navigating an unknown, semi-structured agricultural environment $\mathcal{E}\subset\mathbb{R}^3$ with crops arranged in rows. Although $\mathcal{E}$ is three-dimensional, we model robot motion on the ground plane and represent its configuration in $\mathrm{SE(2)}$. Let $\mathcal{F}_w$ denote a fixed world frame and $\mathcal{F}_b$ the robot's body frame, which moves with the robot. At initialization, the pose of the robot is $\mathbf{x}^w_0 = (\mathbf{p}^w_0, \mathbf{R}^w_0)$, where
$\mathbf{p}^w_0 \in \mathbb{R}^2$ is its planar position and $\mathbf{R}^w_0 \in \mathrm{SO(2)}$ its planar orientation, both expressed in $\mathcal{F}_w$. A navigation task is defined as the tuple $\mathcal{T} = (\mathbf{x}^w_0, \mathbf{p}^w_g)$, where $\mathbf{p}^w_g \in \mathbb{R}^2$ is the goal position in $\mathcal{F}_w$. The goal is sampled in the navigable space of $\mathcal{E}$, up to three crop rows away from $\mathbf{p}^w_0$ and within $\pm 90^\circ$ of $\mathbf{R}^w_0$ to prevent U-turns at the start of the episode. The robot perceives its surroundings only through an egocentric depth camera, which provides an image $\mathbf{I}_t \in \mathbb{R}^{h \times w}$, of $h$ height and and $w$ width every timestep.
We do \emph{not} assume any apriori map of the environment or knowledge of the field geometry such as crop and row spacing, We assume, however, that the robot can estimate its pose $\mathbf{x}^w_t$ at each timestep $t$. This allows the world-frame goal $\mathbf{p}^w_g$ to be expressed in the robot's body frame as $\mathbf{p}^b_g$. The objective is to learn a policy that generates a collision-free path conditioned on the current observation and body-frame goal $\mathbf{p}^b_g$. The task is successful if the robot reaches within is $d = \mid \mid \mathbf{p}^w_t - \mathbf{p}^w_g\mid \mid <1\,$m of the goal, within a time budget of $T$ timesteps.


\section{Approach}
\label{sec:approach}
We present \emph{FlowRow}, a goal-conditioned policy that maps egocentric depth images and a body-frame goal to navigation behaviors for long-horizon navigation in agricultural environments. In this section, we describe the key design choices that enable the policy to generate these behaviors from onboard observations. Our policy doesn't assume a map of the environment or prior knowledge of its geometry. Instead, it acts as a local planner that generates a \emph{short-horizon path} conditioned on the \emph{long-horizon goal} and the robot's observations. An overview of our policy architecture is shown in~\figref{fig:approach}. At each planning timestep $t$, the policy receives the body-frame goal $\mathbf{p}^b_g$ and maintains a history of recent depth images $\mathbf{H}_t$. These inputs are encoded and fused by a transformer to produce a context representation $\mathcal{C}$. Conditioned on $\mathcal{C}$, the flow-matching policy generates $M$ $\mathrm{SE}(2)$ B-Spline control poses. These control poses are used to compute $K$ candidate paths. A path $\tau^*$ is selected from the candidates and tracked by a pure-pursuit controller~\cite{macenski2023rpp} at a constant velocity. At the next timestep, the policy receives a new depth observation, updates $\mathbf{H}_t$ and $\mathbf{p}^b_g$, and generates a new set of candidate paths. This receding-horizon process continues until the robot reaches the world-frame goal $\mathbf{p}^w_g$.

\subsection{Long-Horizon Goal Conditioning}
\label{sec:approach:goal}

As a local planner, FlowRow predicts a short-horizon path within the observable space of the depth image, while the goal $\mathbf{p}^w_g$ may lie tens of meters away. Depending on the robot's distance to the goal, it may lie within or beyond this planning horizon. When beyond the horizon, the goal must guide local planning over multiple planning steps. We encode the goal in the robot's body frame using its range and bearing, and train the policy on both goals within and beyond the planning horizon.

\textbf{Polar Goal Encoder.} We represent the body-frame goal $\mathbf{p}^b_g$ in polar coordinates using its range and bearing. The range to the goal is denoted as $r = \lVert \mathbf{p}^b_g \rVert$  and its bearing as $\beta = \operatorname{atan2}(p^b_{g,y},p^b_{g,x})$. We encode the goal as:

\begin{equation}
\small
\varphi(\mathbf{p}^b_g) =
\left(
\frac{\min(r,r_{\max})}{r_{\max}},
\cos\beta,
\sin\beta
\right),
\end{equation}

where the range is normalized and clipped at $r_{\max}$, while the bearing is preserved independently of the goal distance. We set $r_{\max}=30\,\mathrm{m}$. We represent the bearing using $(\cos\beta,\sin\beta)$ to avoid the discontinuity at $\pm\pi$. This allows the policy to preserve the direction of distant goals without requiring the network to represent arbitrarily large distances. The resulting representation $\varphi(\mathbf{p}^b_g)$ is mapped to a $256$-dimensional goal token using a three-layer MLP.

\textbf{Bi-Horizon Supervision.}
FlowRow predicts a short-horizon path, while the navigation goal may lie much farther away. We therefore train the policy using expert demonstrations with goals sampled at two different horizons; details of the demonstration generation are given in Sec.~\ref{sec:experiments:setup}. For each training sample, we select a demonstrated path segment with arc length $\ell_{\mathrm{seg}} \sim \mathcal{U}(6\,\mathrm{m},10\,\mathrm{m})$. With probability $\alpha=0.15$, we sample the goal within this segment and truncate the segment at the goal, so that FlowRow learns to terminate its prediction when the goal lies within the planning horizon. Otherwise, we place the goal farther along the same demonstration at an additional arc length $\ell \sim \mathcal{U}(0,\ell_{\max})$, with $\ell_{\max}=22\,\mathrm{m}$. In this case, the goal may lie beyond the prediction horizon, so FlowRow learns to generate the appropriate local path rather than move greedily toward the goal. For example, for a goal in another row, the policy can generate continued in-row motion followed by a headland turn instead of a direct lateral path through the crops. This enables FlowRow to learn the field-constrained behaviors required for long-horizon navigation.

\subsection{Context Encoding}
\label{sec:approach:observation}
The goal and depth history form the conditioning input to FlowRow. At each planning timestep, the policy maintains a history of four depth observations in a history buffer, $\mathbf{H}_t = [\mathbf{I}_{t-3},..,\mathbf{I}_t]$. We encode this history using the tokenized depth encoder of SanD-Planner~\cite{wang2026sandplanner}. Each depth image is processed by a shared ResNet-18~\cite{he2016resnet}, truncated before its final stage to preserve spatial information relevant to local obstacle avoidance. The resulting feature maps are reduced to an $8\times12$ spatial grid and projected to an embedding dimension of $D=256$, producing 96 tokens per observation. We augment these tokens with fixed two-dimensional sinusoidal positional embeddings and learned frame-index embeddings to capture spatial and temporal information. The goal $\mathbf{p}^b_g$ is encoded as $\varphi(\mathbf{p}^b_g)$ as described in ~\ref{sec:approach:goal}. The depth and goal tokens are concatenated and processed by a four-layer transformer~\cite{sridhar2023nomad,wang2026sandplanner} to produce the context representation $\mathcal{C}$, used to condition the flow-matching model.

\begin{figure*}[!t]
  \centering
  \includegraphics[width=\linewidth]{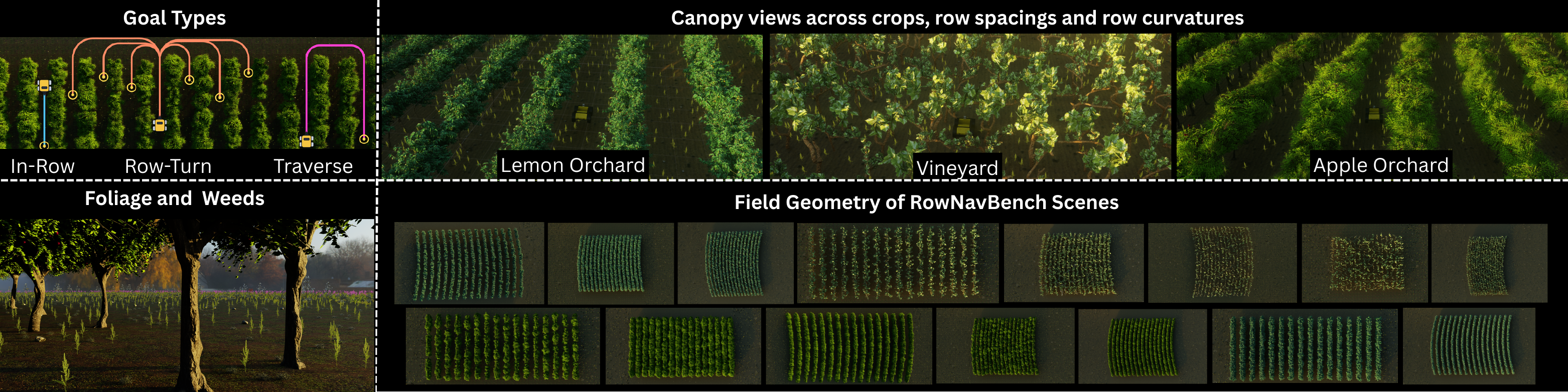}
  \caption{\textbf{RowNavBench Scenes.} Our \emph{RowNavBench} benchmark spans 900 point-goal navigation episodes across 15 scenes and 3 crop types (Lemon Orchard, Vineyard, Apple Orchard), with diverse row lengths, spacing, curvature, crop placement, missing crops, and weed/stone density. We illustrate the episodic goal types, isometric and top-down views of representative scenes from the 15 scenes of RowNavBench.}
  \label{fig:rownavbench}
    \vspace{-0.65cm}
\end{figure*}

\subsection{Action Representation}
\label{sec:approach:action}

Given the context representation $\mathcal{C}$, our flow-matching policy generates actions. We represent the action to be generated as a set of B-spline control poses that compactly parameterize a short-horizon local path. Unlike position-only control points used to parameterize B-splines~\cite{wang2026sandplanner}, we represent each control pose in $\mathrm{SE}(2)$, following related work on Lie-group splines~\cite{lovegrove2013splinefusion,sommer2020cumulative}. This explicitly represents both position and heading along the path, which is important for our wheeled platform. To express the short-horizon path, our policy emits $M=12$ control poses $\mathbf{X}_j \in \mathrm{SE}(2)$, $j=0,\ldots,M-1$, expressed relative to the current robot body frame, each with position $\mathbf{p}_j=(x_j,y_j)^{\top}\in\mathbb{R}^2$ and heading $\theta_j$. We decode the control poses into a continuous path using a cubic cumulative B-spline on $\mathrm{SE}(2)$~\cite{lovegrove2013splinefusion,sommer2020cumulative}. For local spline segment $i$,
\begin{equation}
\begin{aligned}
\mathbf{X}(u)
&=
\mathbf{X}_i
\prod_{k=1}^{3}
\exp\!\left(
\lambda_k(u)\,\mathbf{d}_{i+k}
\right),\\
\mathbf{d}_k
&=
\log\!\left(
\mathbf{X}_{k-1}^{-1}\mathbf{X}_{k}
\right),
\end{aligned}
\label{eq:spline}
\end{equation}
where $\mathbf{d}_k$ denotes the relative motion between consecutive control poses, $\lambda_k(u)$ are the cumulative cubic B-spline basis functions that progressively weight these relative motions, and $u$ parameterizes the local segment. The four poses $\mathbf{X}_i,\ldots,\mathbf{X}_{i+3}$ determine each segment, while clamping the spline ensures that the generated path interpolates the first and last control poses. Although each control pose has three degrees of freedom, we encode it in five dimensions to provide a continuous representation to the flow-matching model. Headland turns can involve large heading changes, including near-$\pi$ rotations, for which direct angle representations are discontinuous at $\pm\pi$. We therefore represent each control pose using bearing and relative heading sine--cosine pairs as
\begin{equation}
\rho_j = \lVert \mathbf{p}_j \rVert,
\qquad
b_j = \operatorname{atan2}(y_j,x_j),
\qquad
\delta_j = \theta_j-b_j .
\label{eq:pose_polar}
\end{equation}
We distinguish $(\rho_j,b_j,\delta_j)$ from the goal range and bearing $(r,\beta)$ defined in Sec.~\ref{sec:approach:goal}. The former parameterize individual control poses along the short-horizon path, whereas $(r,\beta)$ specify the navigation goal that conditions the policy. We encode each control pose as
\begin{equation}
\mathbf{a}_j =
\left(
\frac{2\rho_j}{R_c}-1,\;
\cos b_j,\sin b_j,\;
\cos\delta_j,\sin\delta_j
\right)
\in [-1,1]^5,
\label{eq:encoding}
\end{equation}
where $R_c=10.5\,\mathrm{m}$ is the maximum control-pose range. The resulting representation
$\mathbf{A}=[\mathbf{a}_0,\ldots,\mathbf{a}_{M-1}]
\in[-1,1]^{5\times M}$
is the action our flow-matching policy models. For each training sample, we fit the $M$ control poses to the demonstrated segment of Sec.~\ref{sec:approach:goal} to create the target $\mathbf{A}$ for our flow-matching policy. The first control pose is fixed at the current robot pose so that every generated path starts from the current state. For this pose, $\rho_0=0$ and the bearing is undefined, so we set $b_0=\delta_0=0$ by convention. 

\subsection{Flow-Matching Policy}
\label{sec:approach:flow}

Our action representation parametrizes a local path with SE(2) spline control poses, but multiple valid paths may exist for the same observation and goal. We therefore model a conditional distribution over feasible paths using flow matching. We use conditional flow matching with an independent coupling~\cite{lipman2023fm}, learning a context-conditioned velocity field that transports Gaussian noise to the demonstrated action distribution. Let $\mathbf{z}_1=\mathbf{A}$ denote the encoded action from Sec.~\ref{sec:approach:action}, and let $\mathbf{z}_0\sim\mathcal{N}(\mathbf{0},\mathbf{I})$ be a Gaussian noise sample. For flow time $s\in[0,1]$, we define
\begin{equation}
    \mathbf{z}_s = (1-s)\,\mathbf{z}_0 + s\,\mathbf{z}_1,
    \label{eq:flow_interp}
\end{equation}
with target velocity $\mathbf{z}_1-\mathbf{z}_0$. The velocity network $v_\psi$, parameterized by $\psi$ and conditioned on the context $\mathcal{C}$, is trained using
\begin{equation}
    \mathcal{L}(\psi)
    =
    \mathbb{E}_{s,\mathbf{z}_0,\mathbf{z}_1}
    \left[
    \left\|
    v_\psi(\mathbf{z}_s,s,\mathcal{C})
    -
    (\mathbf{z}_1-\mathbf{z}_0)
    \right\|_2^2
    \right].
    \label{eq:flow_loss}
\end{equation}
We implement $v_\psi$ as a 1-D conditional U-Net over the $M$ control poses, conditioned on $\mathcal{C}$ through cross-attention and on $s$ through an additive embedding. At inference, we sample $\mathbf{z}_0\sim\mathcal{N}(\mathbf{0},\mathbf{I})$ and integrate the learned ODE with $N=10$ Euler steps. The resulting sample is clamped to the valid encoding bounds, decoded into control poses, and used to construct the $\mathrm{SE}(2)$ spline. Repeating this process yields multiple candidate paths to the goal.

\subsection{Path Selection}
\label{sec:approach:selection}

At each planning step, we sample $K=32$ candidate paths in parallel from the flow-matching model. We construct an ESDF from the current depth image and score each candidate path $\tau$ using a clearance cost and a goal-distance cost following Sand-Planner's geometric ranking~\cite{wang2026sandplanner} as:
\begin{equation}
\small
J(\tau)
=
\mathrm{w}_c \, J_{\mathrm{esdf}}(\tau)
+
\mathrm{w}_g
\left\|
\tau_{\mathrm{end}}-\mathbf{p}^{b}_{g}
\right\|^2 .
\label{eq:cost}
\end{equation}
Here, $J_{\mathrm{esdf}}(\tau)$ penalizes paths that pass too close to crops, with greater weight on the portion of the path that is executed before re-planning. We use $\mathrm{w}_c=10^4$ and $\mathrm{w}_g=0.5$, weighing safe obstacle free paths higher than paths which are more goal oriented. The minimum-cost path $\tau^*$ is selected at every planning step and tracked with a pure-pursuit controller on the robot with a constant velocity profile.

\newcolumntype{I}{!{\hspace{0.7em}\vrule\hspace{0.7em}}}
\begin{table*}[!t]
    \centering
    \footnotesize
    \setlength{\tabcolsep}{4pt}
    \renewcommand{\arraystretch}{1.1}
    \begin{tabular*}{\textwidth}{@{\extracolsep{\fill}}cc !{\hspace{1em}} ccI ccccI c@{}}
        \toprule
        \multicolumn{2}{@{}c}{\textbf{\textsc{RowNavBench}}}
          & \multicolumn{2}{cI}{\textsc{Oracle}}
          & \multicolumn{4}{cI}{\textsc{Baselines}}
          & \multirow{2}{*}[-2.3pt]{\shortstack[c]{\textbf{\textsc{{FlowRow}}}}} \\
        \cmidrule(l{.5em}r{1.5em}){3-4}\cmidrule(l{-.4em}r{1.5em}){5-8}
        \textsc{Split} & \textsc{Metrics}
  & A$^*$-Path & Reference-Path
          & NavDP~\cite{cai2025navdp} & SanD-Pl.~\cite{wang2026sandplanner} & SanD-Pl.~[Agri] & SanD-Pl.~[WS-Agri]
          & (Ours) \\
        \midrule
        \cmidrule{1-9}
        \multirow{5}{*}{\shortstack[c]{\textsc{In-Row}\\[3pt] \scriptsize$300$ Episodes}}
          & \met{SR}{\uparrow} & $1.00$ & $1.00$ & $0.98$ & $\mathbf{1.00}$ & $0.93$ & $0.99$ & $0.99$ \\
          & \met{SPL--A$^*$}{\uparrow} & $0.99$ & $0.99$ & $0.98$ & $\mathbf{0.99}$ & $0.93$ & $\mathbf{0.99}$ & $\mathbf{0.99}$ \\
          & \met{SPL--Ref}{\uparrow} & $0.99$ & $0.99$ & $0.97$ & $\mathbf{0.99}$ & $0.93$ & $\mathbf{0.99}$ & $0.98$ \\
          & \met{CTE~[m]}{\downarrow} & $0.24$ & $0.00$ & $0.18$ & $0.11$ & $0.13$ & $0.07$ & $\mathbf{0.03}$ \\
          & \met{HE~[deg]}{\downarrow} & $3.44$ & $0.16$ & $2.89$ & $2.85$ & $1.81$ & $1.24$ & $\mathbf{0.91}$ \\
        \cmidrule{1-9}
        \multirow{5}{*}{\shortstack[c]{\textsc{Row-Turn}\\[3pt] \scriptsize$300$ Episodes}}
          & \met{SR}{\uparrow} & $1.00$ & $1.00$ & $0.00$ & $0.00$ & $0.01$ & $0.24$ & $\mathbf{0.80}$ \\
          & \met{SPL--A$^*$}{\uparrow} & $1.00$ & $0.60$ & $0.00$ & $0.00$ & $0.01$ & $0.19$ & $\mathbf{0.51}$ \\
          & \met{SPL--Ref}{\uparrow} & $1.00$ & $1.00$ & $0.00$ & $0.00$ & $0.01$ & $0.24$ & $\mathbf{0.71}$ \\
          & \met{CTE~[m]}{\downarrow} & $1.00$ & $0.10$ & -- & -- & $1.19$ & $\mathbf{0.89}$ & $1.08$ \\
          & \met{HE~[deg]}{\downarrow} & $53.43$ & $7.69$ & -- & -- & $47.55$ & $41.22$ & $\mathbf{30.82}$\\
        \cmidrule{1-9}
        \multirow{5}{*}{\shortstack[c]{\textsc{Traverse}\\[3pt] \scriptsize$300$ Episodes}}
          & \met{SR}{\uparrow} & $0.99$ & $1.00$ & $0.00$ & $0.01$ & $0.00$ & $0.09$ & $\mathbf{0.82}$ \\
          & \met{SPL--A$^*$}{\uparrow} & $0.99$ & $0.84$ & $0.00$ & $0.01$ & $0.00$ & $0.08$ & $\mathbf{0.69}$ \\
          & \met{SPL--Ref}{\uparrow} & $0.99$ & $1.00$ & $0.00$ & $0.01$ & $0.00$ & $0.09$ & $\mathbf{0.78}$ \\
          & \met{CTE~[m]}{\downarrow} & $0.96$ & $0.06$ & -- & $1.21$ & -- & $\mathbf{0.55}$ & $0.65$ \\
          & \met{HE~[deg]}{\downarrow} & $34.76$ & $4.86$ & -- & $45.32$ & -- & $27.27$ & $\mathbf{20.12}$ \\
        \cmidrule{1-9}
        \multirow{5}{*}{\shortstack[c]{\textsc{All}\\[3pt] \scriptsize$900$ Episodes}}
          & \met{SR}{\uparrow} & $0.99$ & $1.00$ & $0.32$ & $0.34$ & $0.31$ & $0.44$ & $\mathbf{0.87}$ \\
          & \met{SPL--A$^*$}{\uparrow} & $0.99$ & $0.81$ & $0.32$ & $0.34$ & $0.31$ & $0.42$ & $\mathbf{0.73}$ \\
          & \met{SPL--Ref}{\uparrow} & $0.99$ & $0.99$ & $0.32$ & $0.33$ & $0.31$ & $0.44$ & $\mathbf{0.82}$ \\
          & \met{CTE~[m]}{\downarrow} & $0.73$ & $0.05$ & $0.18$ & $\mathbf{0.14}$ & $0.15$ & $0.26$ & $0.55$ \\
          & \met{HE~[deg]}{\downarrow} & $30.53$ & $4.24$ & $3.02$ & $3.93$ & $\mathbf{2.69}$ & $10.40$ & $16.10$ \\

        \bottomrule
    \end{tabular*}
    \caption[Benchmarking on RowNavBench]{We present the evaluation of our FlowRow policies compared to baselines on the PointGoal task on our benchmark \textit{RowNavBench}. We report the success rate~(SR), success weighted by path length~(SPL), and the cross-track error~(CTE) and heading error~(HE) for successful epiosdes, as RMSE against the reference path for the three splits of the benchmark, i.e., \textsc{In-Row}, \textsc{Row-Turn}, and \textsc{Traverse}, comprising of $300$ episodes each. We finally present the metrics for the \textsc{All} aggregate over $900$ episodes.}
    \label{tab:RowNavBench_results}
\end{table*}

\section{Experimental Evaluation}
\label{sec:experiments}

We design our experimental evaluation to assess the effectiveness, generalization, and robustness of \textit{FlowRow} across simulated and real-world agricultural environments with varied field geometry. Specifically, our experiments address four key aspects: (i) performance on \textit{RowNavBench} compared to existing local planning approaches, (ii) zero-shot deployment of the simulation-trained policy in a real-world agricultural environment, (iii) robustness to localization errors compared to a controller tracking a pre-planned global reference path, and (iv) the role of our goal representation, action representation, and training supervision in enabling long-horizon navigation through systematic ablations.

\begin{figure}[!h]
  \centering
  \includegraphics[width=0.9\linewidth]{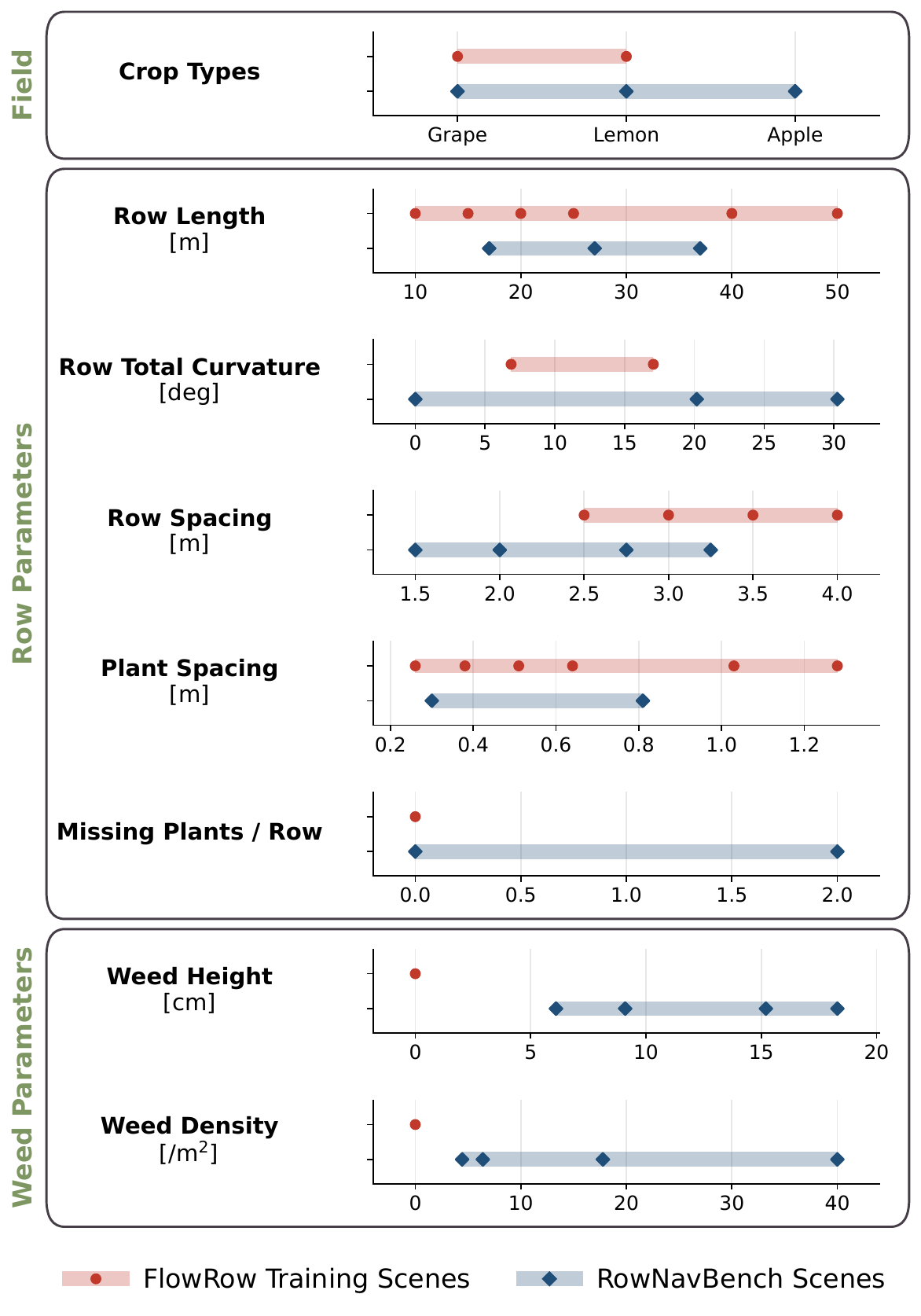}
  \caption{\textbf{Benchmark Statistics.} We compare crop-field parameter variation across \textit{RowNavBench} scenes against those used to collect FlowRow’s training demonstrations. FlowRow's training data is collected from weedless Lemon orchards and Vineyards, with Apple Orchards held out as an out-of-distribution set.}
  \label{fig:train_vs_bench}
    \vspace{-0.7cm}
\end{figure}

\subsection{Experimental Setup}
\label{sec:experiments:setup}

\textbf{Benchmark.}
We introduce RowNavBench, a simulation benchmark for long-horizon, goal-directed navigation in semi-structured agricultural environments. It comprises $900$ episodes, evenly split into $300$ \emph{in-row}, $300$ \emph{row-turn}, and $300$ \emph{traverse} episodes, illustrated in Fig.~\ref{fig:rownavbench}. \emph{In-row} episodes place the goal ahead within the current row. \emph{Row-turn} episodes place it up to three rows to the left or right, requiring the robot to exit the current row, skip rows when needed, and enter the target row. \emph{Traverse} episodes additionally combine in-row driving, headland turning, and row skipping to reach a target row up to three rows away. Episodes span $15$ simulation environments generated with AgriGen~\cite{bajpai2026agrigen}, with $3$ crop types, $3$ row lengths, $3$ row curvatures, $4$ row spacings, $2$ plant spacings, $2$ missing-plant settings, $4$ weed heights, and $4$ weed densities. Each crop type uses $10$ artist-designed crop models with randomized yaw, while rocks, grass, and weeds provide additional variation. RowNavBench provides both an A$^\star$ shortest path and an agricultural reference path that remains centered in-row and turns only at the headland. The goal is expressed in the robot body frame, and all policies use the same Pure Pursuit Controller~\cite{macenski2023rpp}, without an internal costmap, under a constant velocity profile.

\textbf{Training Demonstrations.}
We generate training data in AgriGen~\cite{bajpai2026agrigen} using field geometries disjoint from RowNavBench to avoid data leakage. The training and benchmark distributions are shown in~\figref{fig:train_vs_bench}. We collect $178$ navigation demonstrations in weedless lemon and grape scenes. In $114$ demonstrations, the robot follows boustrophedon traversals along reference paths computed from the field geometry. Turns are initiated beyond the row ends with a safety margin from the crops. We additionally record $64$ recovery demonstrations, in which the robot is initialized off-track and driven back toward the reference path.

\textbf{Evaluation Metrics.}
A navigation episode succeeds if the robot reaches the goal region
$\|\mathbf{p}^w_t-\mathbf{p}^w_g\|_2 < d$ within the timestep budget $T$, without colliding or leaving the field. Over a split of $N$ episodes, we report success rate as
\vspace{-0.5em}
\begin{equation}
\small
    \mathrm{SR} = \frac{1}{N}\sum_{i=1}^{N} S_i,
    \qquad S_i \in \{0,1\}.
    \label{eq:sr}
    \vspace{-0.5em}
\end{equation}
To measure path quality, we report success weighted by path length (SPL), cross-track error (CTE), and heading error (HE). SPL compares the distance traveled by the robot $\ell_i$ with the corresponding reference path length $\ell_i^\star$, as :
\vspace{-0.5em}
\begin{equation}
\small
    \mathrm{SPL}
    = \frac{1}{N}\sum_{i=1}^{N}
    S_i
    \min\!\left(
        \frac{\max(\ell_i^\star-d,\,0)}{\ell_i},
        1
    \right).
    \label{eq:spl}
\end{equation}
%
Subtracting $d$ accounts for termination upon entering the goal region. We report \textbf{SPL-A$^\star$} using the A$^\star$ shortest path and \textbf{SPL-Ref} using the agricultural reference path. We measure path tracking quality relative to the agricultural reference path $\mathcal{P}$ using CTE and HE. For each robot pose $\mathbf{x}^w_t=(\mathbf{p}^w_t,\mathbf{R}^w_t)$, we project $\mathbf{p}^w_t$ onto $\mathcal{P}$ to obtain the signed lateral error $e_t$ and reference heading $\bar{\psi}_t$. For a navigation episode with $n$ recorded poses,
\begin{equation}
\small
    \mathrm{CTE}
    = \sqrt{\frac{1}{n}\sum_{t=1}^{n} e_t^2},
    \qquad
    \mathrm{HE}
    = \sqrt{\frac{1}{n}\sum_{t=1}^{n}
    \left[\mathrm{wrap}(\psi_t-\bar{\psi}_t)\right]^2}.
    \label{eq:cte}
\end{equation}
We report the mean per-episode $\mathrm{CTE}$ and $\mathrm{HE}$ over successful navigation episodes.

\textbf{Baselines.}
We compare FlowRow against two privileged planners and four learned baselines. \emph{Oracle-A$^\star$-Path} and \emph{Oracle-Reference-Path} assume full knowledge of the field and exact robot pose, and track the A$^\star$ shortest path and agricultural reference path, respectively. They serve as oracle baselines and define the reference lengths used by SPL-A$^\star$ and SPL-Ref. \emph{NavDP}~\cite{cai2025navdp} is a diffusion policy for point-goal navigation using RGB-D perception, evaluated using its released weights. \emph{SanD-Pl.}~\cite{wang2026sandplanner} is evaluated using its released checkpoint. Since this checkpoint was trained on indoor environments, we additionally evaluate \emph{SanD-Pl.[Agri]}, trained from scratch on our agricultural data using the authors' code and hyperparameters, and \emph{SanD-Pl.[WS-Agri]}, warm-started from the released checkpoint and trained on our data for the full schedule. All learned baselines use the same observations, and control interface as FlowRow.
\subsection{Performance Evaluation on RowNavBench}
\label{sec:exp:rownavbench}
Our first experiment evaluates FlowRow on RowNavBench against the privileged oracle planners and learned baselines introduced above. As shown in Table~\ref{tab:RowNavBench_results}, we report the metrics defined in the previous section over the three navigation splits and the overall aggregate. FlowRow achieves an overall success rate of \textbf{87\%}, substantially outperforming \emph{NavDP} $(\mathbf{32\%})$ and \emph{SanD-Pl.} $(\mathbf{34\%})$. Training \emph{SanD-Pl.[Agri]} from scratch on the same $178$ agricultural trajectories used by FlowRow, while following the authors' training recipe, yields a success rate of $\mathbf{31\%}$. Warm-starting \emph{SanD-Pl.[WS-Agri]} from the released \emph{SanD-Pl.} checkpoint and training on the same agricultural data improves performance to $\mathbf{44\%}$. In comparison, FlowRow achieves $\mathbf{87\%}$ success using the same $178$ training trajectories. These results indicate that FlowRow provides higher navigation success and better data efficiency for goal-directed navigation in semi-structured agricultural environments.
\begin{figure}
  \centering
  \includegraphics[width=0.99\linewidth]{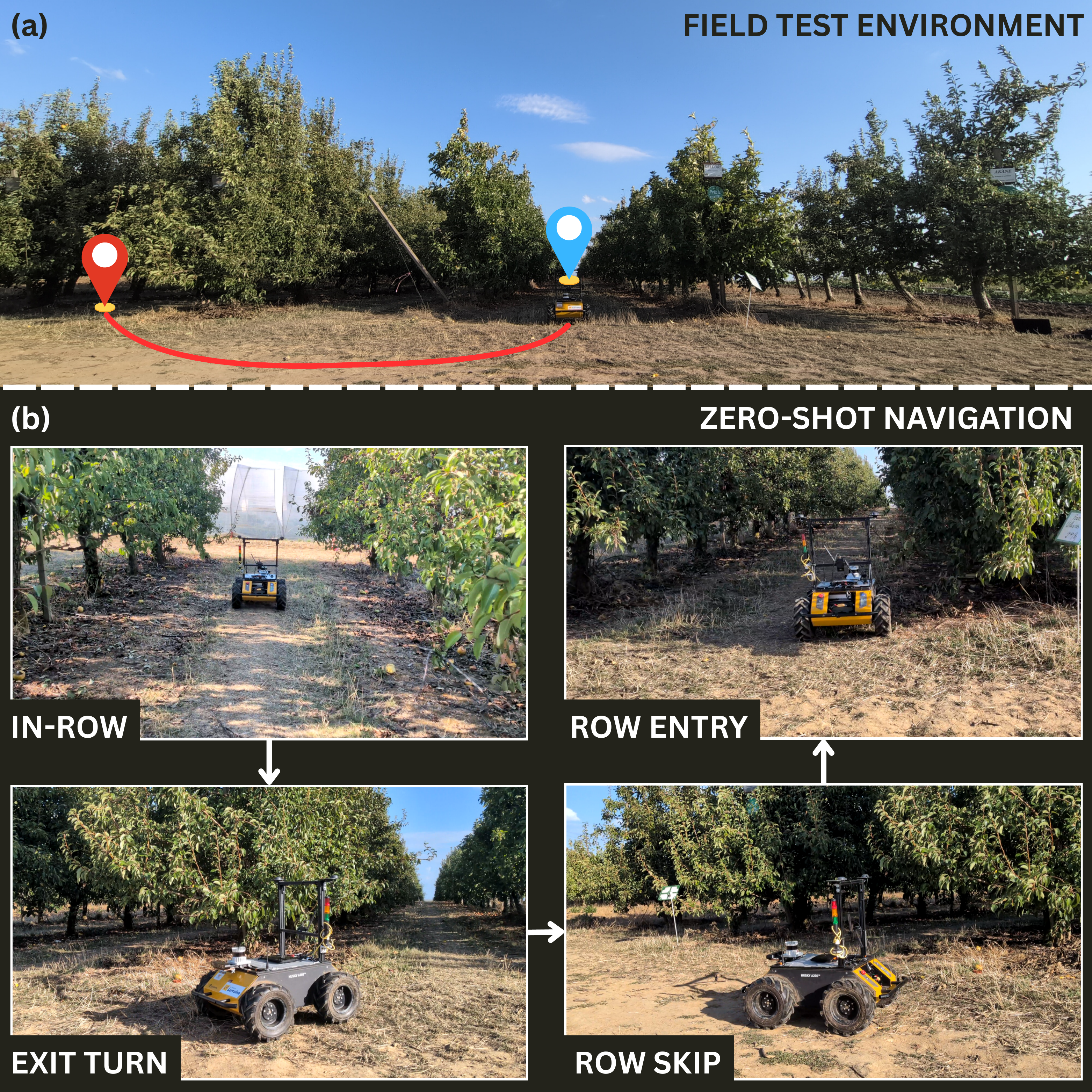}
  \caption{We deploy FlowRow, trained entirely on simulated data, in a zero-shot manner in a real apple orchard. (a) The field test environment consists of parallel rows of apple trees with varied outgrowth and clutter. The robot is initialized~(blue marker) in one row and tasked with reaching the goal~(red marker) in a row separated by one intervening row. (b) Snapshots show the robot’s progress as it drives in-row, turns at the headland, skips one row, and enters the target row to reach the goal.
}
  \label{fig:field_test}
\end{figure}
\vspace{-0.31em}
\subsection{Zero-Shot Real-World Deployment}
\label{sec:exp:real}
We deploy \textit{FlowRow} zero-shot in real-world orchards after training exclusively on simulation demonstrations. We conduct $8$ real-world experiments in apple and pear orchards with row spacings of $3.8$\,m and $3.2$\,m, respectively, under weeds, fallen fruit, vegetation clutter, and uneven terrain, as shown in~\figref{fig:field_test}. FlowRow runs on a Husky A200 robot onboard an NVIDIA AGX Orin at $1.5$\,Hz and receives $\mathbf{I}_t \in \mathbb{R}^{180 \times 240}$ depth images from a ZED 2i stereo camera. RTK-GNSS provides $\mathbf{x}^w_t$, which we use to transform the fixed UTM goal $\mathbf{p}^w_g$ into the body frame as $\mathbf{p}^b_g$. All $3$ in-row experiments succeed, while $3$ of $5$ traverse experiments complete without intervention. We additionally test out-of-distribution conditions, including obstacles occluding the camera view and a traverse spanning five rows. In the latter case, a thin signpost near the row entrance is not reliably perceived by the depth sensor, requiring a brief intervention before FlowRow successfully completes the out-of-distribution traverse.

\begin{figure}[t]
  \centering
  \includegraphics[width=0.99\linewidth]{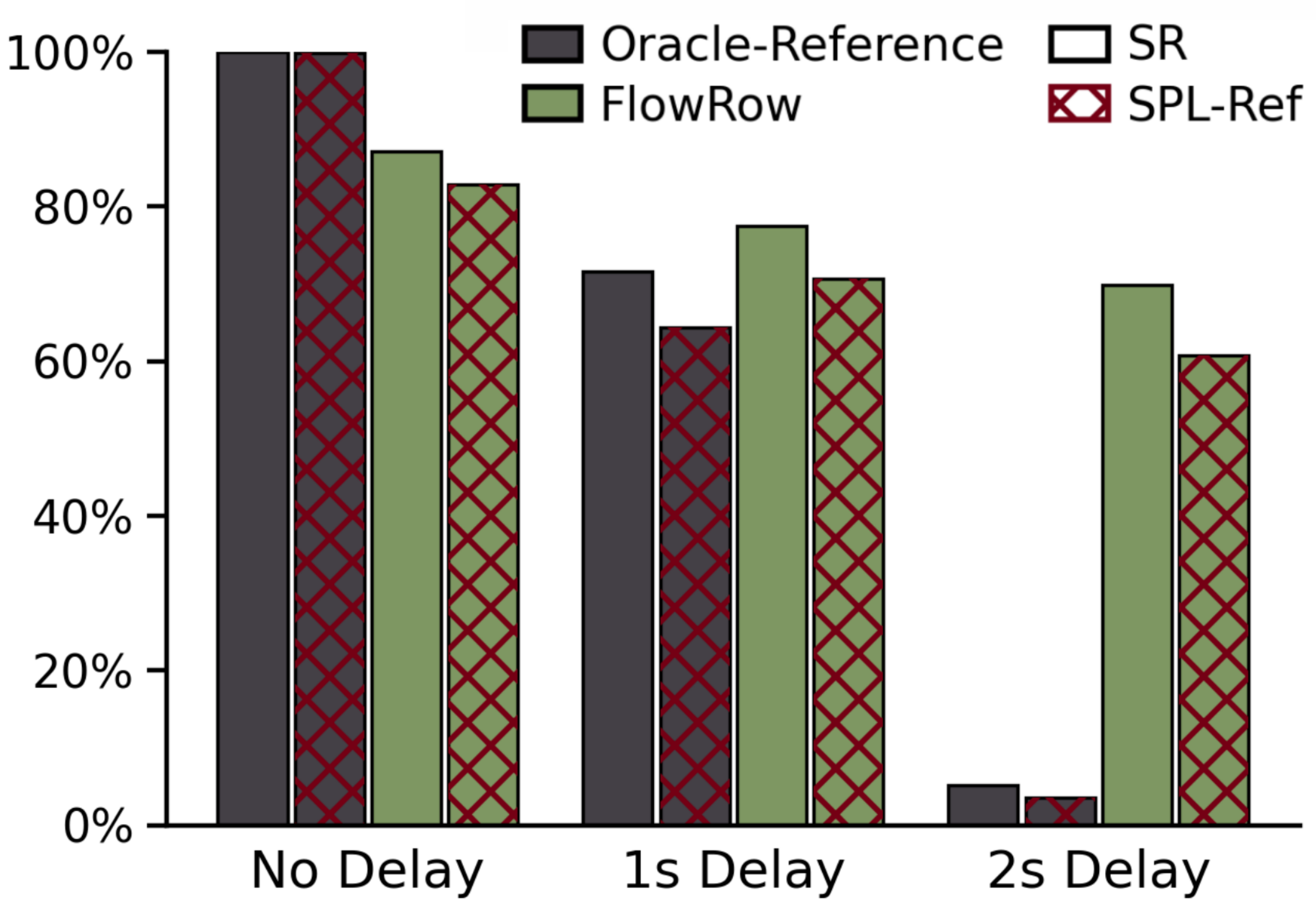}
  \caption{We evaluate the robustness of FlowRow against the reference oracle under delayed localization on RowNavBench. We compare three scenarios, perfect localization, and localization delayed by $1$s and $2$s. We report SR and SPL-Ref metrics and observe that FlowRow is more robust to delays compared to the oracle.
}
  \label{fig:loc_delay}
\end{figure}

\subsection{Robustness to Localization Errors}
\label{sec:exp:locdelay}
We evaluate robustness to delayed localization, a localization error representative of real-world errors, by providing each planner with a pose estimate delayed by $t$ timesteps, while keeping depth observations and metric computation synchronized with the true pose. We evaluate delays of $1$\,s  and $2$\,s over all $900$ RowNavBench episodes, and report the success rates in ~\figref{fig:loc_delay}. We apply the same delayed pose to \emph{FlowRow} and \emph{Oracle-Reference-Path}. The oracle requires localization to track its world-frame reference path, whereas FlowRow uses localization only to transform the goal into the body frame and its local obstacle avoidance remains conditioned on current depth observations. \figref{fig:loc_delay} shows that although \emph{Oracle-Reference-Path} outperforms FlowRow as expected under exact localization ($100$\% vs $87$\% SR), its performance drops significantly as localization error increases, while FlowRow remains comparatively stable.

\subsection{Effect of Bi-Horizon Supervision}
\label{sec:exp:representation}
We study how training FlowRow's flow matching model with different supervision horizons affects long-horizon navigation. As shown in Tab.~\ref{tab:ablation_supervision}, we compare three supervision regimes: \emph{Fixed-Horizon} supervision trains the model with paths of approximately $8$\,m arclength ($6$--$10$\,m), placing the goal at the end of the path. \emph{Short-Horizon} supervision samples the path uniformly from $0.46$ to $10$\,m and places the goal at the end of the path, increasing the model's ability to generate short paths if the goal is near the robot. \emph{Bi-Horizon} supervision (\ref{sec:approach:goal}) trains FlowRow on paths of approximately $8$\,m arclength but samples the goal beyond $8$\,m with a $85\%$ probability and within $8$\,m with a $15\%$ probability, decoupling action and goal horizons. We evaluate the effect of supervision on two different combinations of action and goal representations. Our polar goal and $\mathrm{SE}(2)$ spline representation (\ref{sec:approach:goal}, \ref{sec:approach:action}), and a cartesian goal and $\mathbb{R}^3$ spline similar to~\cite{wang2026sandplanner} We evaluate their performance on the longest 150 \textsc{Traverse} episodes of \textit{RowNavBench}. 

Our results indicate that supervision horizon is the main driver of long-horizon performance. Policies trained with Fixed-Horizon supervision yield near-zero success and frequent collisions for both representations. Short-Horizon improves performance, with a larger gain for polar-$SE(2)$ representation, while Bi-Horizon reaches $87.3\%$ and $84.0\%$ success for the polar-$\mathrm{SE}(2)$ and $\mathbb{R}^3$ configurations, with 8 and 22 collisions and 11 and 1 timeouts (T.O), respectively. We find that Bi-Horizon supervision substantially increases long-horizon success and reduces sensitivity to the choice of action representation. 
In summary, our evaluation supports the claims of this work. FlowRow achieves strong performance on RowNavBench, enables zero-shot deployment in agricultural environments, and exhibits robustness to localization errors. We further examine the effect of supervision horizon on the representational design choices of FlowRow.

\begin{table}[t]
  \centering
  \setlength{\tabcolsep}{4pt}
  \resizebox{\columnwidth}{!}{%
  \begin{tabular}{ll cc cc}
    \toprule
    \multicolumn{2}{c}{Recipe} & \multicolumn{2}{c}{Metrics} & \multicolumn{2}{c}{Failures} \\
    \cmidrule(lr){1-2} \cmidrule(lr){3-4} \cmidrule(lr){5-6}
    Supervision & Action + goal
      & SR $\uparrow$ & SPL-Ref $\uparrow$
      & Coll. $\downarrow$ & T.O. $\downarrow$ \\
    \midrule
    \multirow{2}{*}{\textsc{fixed-horizon}} & SE(2) + polar & 0.02 & 0.02 & $\mathbf{142}$ & 5 \\
     & $\mathbb{R}^3$ + cartesian & $\mathbf{0.04}$ & $\mathbf{0.04}$ & 144 & $\mathbf{0}$ \\
    \midrule
    \multirow{2}{*}{\textsc{short-horizon}} & SE(2) + polar & $\mathbf{0.35}$ & $\mathbf{0.34}$ & $\mathbf{97}$ & 0 \\
     & $\mathbb{R}^3$ + cartesian & 0.11 & 0.11 & 132 & 0 \\
    \midrule
    \multirow{2}{*}{\textsc{bi-horizon}} & SE(2) + polar & $\mathbf{0.87}$ & $\mathbf{0.85}$ & $\mathbf{8}$ & 11 \\
     & $\mathbb{R}^3$ + cartesian & 0.84 & 0.83 & 22 & $\mathbf{1}$ \\
    \bottomrule
  \end{tabular}%
  }
  \caption[Ablation of supervision]{Supervision horizon ablation on the longest \textsc{Traverse} episodes of \textit{RowNavBench}. We compare the effect of supervision horizon on FlowRow's polar goal+\,SE(2)\, representation vs a cartesian goal $\,+\, \mathbb{R}^3$ representation. Bi-horizon supervision substantially raises success across both representations. }
    \label{tab:ablation_supervision}
        \vspace{-0.65cm}
\end{table}

\section{Conclusion}
\label{sec:conclusion}
In this work, we proposed \textit{FlowRow}, a long-horizon goal-conditioned flow matching policy for navigation in semi-structured agricultural environments. We introduce \textit{RowNavBench}, a benchmark for navigation in agricultural environments. We introduce a novel long-horizon goal conditioning mechanism to navigate long horizon environments with a local planner by training with goals beyond the training horizon. We also introduced a novel action space for flow matching policies as $\mathrm{SE}(2)$ spline control poses. Our experimental results compare the performance of our approach to state-of-the art generative imitation learning baselines and show that our approach achieves 87\% success on this task. We further ablate our representation choices and design decisions. We additionally show that FlowRow is robust to localization errors. Finally, we show zero-shot deployement of FlowRow in a real world agricultural environment.
Overall, our findings demonstrate how long-horizon flow matching policies can be useful in agricultural navigation. Future work will conduct extensive evaluation of the sim2real generalization of this approach.


\bibliographystyle{plain_abbrv}

\bibliography{glorified,new}

\IfFileExists{./certificate/certificate.tex}{
\subfile{./certificate/certificate.tex}
}{}
\end{document}

%% file: stachnisslab-latex.tex
\usepackage{graphics}           
\usepackage{times}              
\usepackage{amsmath}            
\usepackage{amssymb}            
\usepackage{graphicx}
\usepackage{algorithm}
\usepackage[noend]{algpseudocode}
\usepackage{booktabs}
\usepackage{color}
\usepackage{listings}
\usepackage{subfiles}
\usepackage{hyperref}
\usepackage[nocompress]{cite} 
\definecolor{instructioncolor}{rgb}{.5,.5,.5}

\usepackage[font=small]{caption}

\def\figref#1{Fig.~\ref{#1}}

\def\eqref#1{Eq.~(\ref{#1})}

\makeatletter
\usepackage{xspace}
\DeclareRobustCommand\onedot{\futurelet\@let@token\@onedot}
\def\@onedot{\ifx\@let@token.\else.\null\fi\xspace}

\def\etal{{et al}\onedot}
\makeatother

\def\etalcite#1{\etal~\cite{#1}}

\usepackage{array}
\newcolumntype{L}[1]{>{\raggedright\let\newline\\\arraybackslash\hspace{0pt}}m{#1}}
\newcolumntype{C}[1]{>{\centering\let\newline\\\arraybackslash\hspace{0pt}}m{#1}}
\newcolumntype{R}[1]{>{\raggedleft\let\newline\\\arraybackslash\hspace{0pt}}m{#1}}

%% file: glorified.bib
@STRING{arxiv                  = {arXiv preprint} }

@STRING{bmvc                   = {Proc.~of British Machine Vision Conf. (BMVC)} }

@STRING{cvpr                   = {Proc.~of the IEEE/CVF Conf.~on Computer Vision and Pattern Recognition (CVPR)} }

@STRING{ecmr                   = {Proc.~of the Europ.~Conf.~on Mobile Robotics (ECMR)} }

@STRING{iclr                   = {Proc.~of the Intl.~Conf.~on Learning Representations (ICLR)}}

@STRING{icra                   = {Proc.~of the IEEE Intl.~Conf.~on Robotics \& Automation (ICRA)} }

@STRING{iros                   = {Proc.~of the IEEE/RSJ Intl.~Conf.~on Intelligent Robots and Systems (IROS)} }

@STRING{jfr                    = {Journal of Field Robotics (JFR)} }

@STRING{ral                    = {IEEE Robotics and Automation Letters (RA-L)} }

@STRING{rss                    = {Proc.~of Robotics: Science and Systems (RSS)} }


%% file: new.bib
@ARTICLE{kim2025pagnav,
  author={{Kim, Kitae and Deb, Aarya and Cappelleri, David J.}},
  journal=ral, 
  title={{P-AgNav: Range View-Based Autonomous Navigation System for Cornfields}}, 
  year={2025},
  volume={10},
  number={4},
}

@article{wang2026sandplanner,
  title={{SanD-Planner: Sample-Efficient Diffusion Planner in B-Spline Space for Robust Local Navigation}},
  author={Wang, Jincheng and Bao, Lingfan and Yang, Tong and Plasencia, Diego Martinez and Jiao, Jianhao and Kanoulas, Dimitrios},
  booktitle=rss,
  year={2026}
}

@article{cai2025navdp,
  title={Navdp: Learning sim-to-real navigation diffusion policy with privileged information guidance},
  author={Cai, Wenzhe and Peng, Jiaqi and Yang, Yuqiang and Zhang, Yujian and Wei, Meng and Wang, Hanqing and Chen, Yilun and Wang, Tai and Pang, Jiangmiao},
  journal={arXiv preprint arXiv:2505.08712},
  year={2025}
}

@INPROCEEDINGS{sivakumar2021rss, 
    author= {Arun Narenthiran Sivakumar AND Sahil Modi AND Mateus Valverde Gasparino AND Che Ellis AND Andres Eduardo {Baquero Velasquez} AND Girish Chowdhary AND Saurabh Gupta}, 
    title= {{Learned Visual Navigation for Under-Canopy Agricultural Robots}}, 
    booktitle=rss, 
    year= {2021}, 
}

@article{sivakumar2024rss,
  title={Lessons from deploying CropFollow++: under-canopy agricultural navigation with keypoints},
  author={Sivakumar, Arun N and Gasparino, Mateus V and McGuire, Michael and Higuti, Vitor AH and Akcal, M Ugur and Chowdhary, Girish},
  booktitle=rss,
  year={2024}
}

@inproceedings{chi2023diffusionpolicy,
	title={Diffusion Policy: Visuomotor Policy Learning via Action Diffusion},
	author={Chi, Cheng and Feng, Siyuan and Du, Yilun and Xu, Zhenjia and Cousineau, Eric and Burchfiel, Benjamin and Song, Shuran},
	booktitle=rss,
	year={2023}
}

@inproceedings{sridhar2023nomad,
    author  = {Ajay Sridhar and Dhruv Shah and Catherine Glossop and Sergey Levine},
    title   = {{NoMaD: Goal Masked Diffusion Policies for Navigation and Exploration}},
    booktitle = icra,
    year    = {2024},
}

@inproceedings{gode2025flownav,
  author={Gode, Samiran and Nayak, Abhijeet and Oliveira, Débora N.P. and Krawez, Michael and Schmid, Cordelia and Burgard, Wolfram},
  booktitle=iros, 
  title={FlowNav: Combining Flow Matching and Depth Priors for Efficient Navigation}, 
  year={2025},
}

@inproceedings{lipman2023fm,
title={Flow Matching for Generative Modeling},
author={Yaron Lipman and Ricky T. Q. Chen and Heli Ben-Hamu and Maximilian Nickel and Matthew Le},
booktitle=iclr,
year={2023},
}

@inproceedings{he2016resnet,
	author={Kaiming He and Xiangyu Zhang and Shaoqing Ren and Jian Sun},
	title={Deep Residual Learning for Image Recognition},
	booktitle=cvpr,
	year={2016}
}

@article{macenski2023rpp,
      title={Regulated Pure Pursuit for Robot Path Tracking}, 
      author={Steve Macenski and Shrijit Singh and Francisco Martin and Jonatan Gines},
      year={2023},
      journal={arXiv preprint, arXiv:2305.20026},
}

@inproceedings{lovegrove2013splinefusion,
  title={Spline Fusion: A continuous-time representation for visual-inertial fusion with application to rolling shutter cameras.},
  author={Lovegrove, Steven and Patron-Perez, Alonso and Sibley, Gabe},
  booktitle={BMVC},
  year={2013}
}

@inproceedings{sommer2020cumulative,
   title={Efficient Derivative Computation for Cumulative B-Splines on Lie Groups},
   booktitle=cvpr,
   author={Sommer, Christiane and Usenko, Vladyslav and Schubert, David and Demmel, Nikolaus and Cremers, Daniel},
   year={2020},
}

@article{oksanen2009coverage,
author = {Oksanen, Timo and Visala, Arto},
title = {Coverage path planning algorithms for agricultural field machines},
journal = jfr,
volume = {26},
number = {8},
pages = {651-668},
year = {2009}
}

@inproceedings{ahmadi2020visual,
  title={Visual servoing-based navigation for monitoring row-crop fields},
  author={Ahmadi, Alireza and Nardi, Lorenzo and Chebrolu, Nived and Stachniss, Cyrill},
  booktitle=icra,
  year={2020},
}

@inproceedings{ahmadi2022towards,
  title={Towards autonomous visual navigation in arable fields},
  author={Ahmadi, Alireza and Halstead, Michael and McCool, Chris},
  booktitle=iros,
  year={2022},
}

@inproceedings{bertoglio2023ecmr,
  author={Bertoglio, Riccardo and Carini, Veronica and Arrigoni, Stefano and Matteucci, Matteo},
  booktitle=ecmr, 
  title={A Map-Free LiDAR-Based System for Autonomous Navigation in Vineyards}, 
  year={2023},
}

@inproceedings{liu2023visioniros,
  title={Vision-based vineyard navigation solution with automatic annotation},
  author={Liu, Ertai and Monica, Josephine and Gold, Kaitlin and Cadle-Davidson, Lance and Combs, David and Jiang, Yu},
  booktitle=iros,
  year={2023},
}

@article{desilva2024jfr,
author = {de Silva, Rajitha and Cielniak, Grzegorz and Wang, Gang and Gao, Junfeng},
title = {Deep learning-based crop row detection for infield navigation of agri-robots},
journal = jfr,
volume = {41},
number = {7},
pages = {2299-2321},
year = {2024}
}

@article{sivakumar2024learningturndiffusionimitation,
      title={Learning to Turn: Diffusion Imitation for Robust Row Turning in Under-Canopy Robots}, 
      author={Arun N. Sivakumar and Pranay Thangeda and Yixiao Fang and Mateus V. Gasparino and Jose Cuaran and Melkior Ornik and Girish Chowdhary},
      year={2024},
      journal={arXiv preprint, arXiv:2408.03059},
}

@misc{bajpai2026agrigen,
  author={Bajpai, Utkarsh and Tleiji, Serge and Pradalier, Cédric and Aravecchia, Stéphanie },
  title={{AgriGen: Large-Scale Scene Generation Framework for Photorealistic Agricultural Robotics Simulation}},
  booktitle={ICRA 2026 Workshop on Agricultural Robotics},
  url={https://openreview.net/forum?id=VWDd4ueN2T},
  year={2026},
}
